\documentclass[letterpaper, 10 pt, conference]{ieeeconf}  % Comment this line out
\IEEEoverridecommandlockouts                              % This command is only
\usepackage{graphicx}
\usepackage{url}
\usepackage{geometry}
\usepackage{listings}
\usepackage{color}

\definecolor{dkgreen}{rgb}{0,0.6,0}
\definecolor{mauve}{rgb}{0.58,0,0.82}

\title{\LARGE \bf
A Genetic Algorithm Based Approach for Satellite Autonomy
}

\author{ \parbox{3 in}{\centering Sidhdharth Sikka\\
        Manifold Computing\\
        {\tt\small sidhsikka1998@g.ucla.edu}}
        \parbox{3 in}{ \centering Harshvardhan Sikka
        \\
        Georgia Institute of Technology \\
        Manifold Computing\\
        OpenMined\\
        {\tt\small harsh@manifoldcomputing.com}}
}

\begin{document}

\maketitle
\thispagestyle{empty}
\pagestyle{empty}

%%%%%%%%%%%%%%%%%%%%%%%%%%%%%%%%%%%%%%%%%%%%%%%%%%%%%%%%%%%%%%%%%%%%%%%%%%%%%%%%
\begin{abstract}

Autonomous spacecraft maneuver planning using an evolutionary algorithmic approach is investigated. Simulated spacecraft were placed into four different initial orbits. Each was allowed a string of thirty delta-v impulse maneuvers in six cartesian directions, the positive and negative x, y and z directions. The goal of the spacecraft maneuver string was to, starting from some non-polar starting orbit, place the spacecraft into a polar, low eccentricity orbit. A genetic algorithm was implemented, using a mating, fitness, mutation and crossover scheme for impulse strings. The genetic algorithm was successfully able to produce this result for all the starting orbits. Performance and future work is also discussed.
\end{abstract}

%%%%%%%%%%%%%%%%%%%%%%%%%%%%%%%%%%%%%%%%%%%%%%%%%%%%%%%%%%%%%%%%%%%%%%%%%%%%%%%%
\section{INTRODUCTION}

Applications of orbital technology grow as the cost of producing satellites and launching them decrease. For the first time in history, diminished launch and development costs are making it possible to create distributed satellite systems. These systems have numerous operational advantages over monolithic satellite design, including but not limited to robustness to the consequences of failure in individual satellites, decreased cost of implementation and replacement due to the size of the units, and being in multiple locations in space. Increases in computing power and miniaturization of the associated platforms has also increased the computational resources available on-board small satellites. This allows for the use of distributed satellite systems in previously intractable problem domains, including planet-wide monitoring for tasks like gathering climate data and tracking wildlife migratory patterns\cite{araguz2018applying}. Additionally, these advances have also opened up possibilities for coordination and autonomy in distributed satellite systems. If a group of satellites were to coordinate, they could accomplish objectives like flying in formation, assembling into more complex structures, retrieval and repair, etc. A short history of the prior research and application of satellite autonomy and distributed coordination (Satellite ADC) is presented below. 

\subsection{History of Autonomy and Distributed Space Systems}
Autonomy in single satellites and spacecraft has been implemented in older missions than autonomy among distributed systems of spacecraft. Implementations of autonomous spacecraft date back to 1997, when the Mars Pathfinder mission, in particular the rover \textit{Sojourner}, employed simple autonomy features including simple landmark based navigation. A year later, \textit{Deep Space 1} was performing autonomous task allocation\cite{bernard1999spacecraft} and image based navigation\cite{bhaskaran1998orbit}, and three years later in 2000, NASA's \textit{Earth Observing 1} was autonomously imaging areas of interest on the Earth\cite{r_j_doyle_1998}. The first major mission successfully demonstrating a form of distributed autonomy was the NASA NODES mission. This mission deployed from the ISS in 2016 and was a demonstration of task allocation among multiple satellites using a coordination scheme called negotiation. Negotiation is when satellites establish a connection with each other, then allocate tasks among each other based on orbital parameters, telemetry and resource information\cite{hanson2016nodes}. The mission was successful in accomplishing its objectives. Currently, NASA has introduced a distributed autonomous mission paradigm called Autonomous Nano-Technology Swarm (ANTS) and have several missions in the planning stage following this paradigm including the Saturn Autonomous Ring Array (SARA), a swarm of small satellites that will closely observe the rings of Saturn\cite{araguz2018applying}.

\subsection{Related Work}
Though the implementations of distributed, autonomous space systems have been few and far in between, exploration of the problem has been a popular topic of academic study around the world. This research has broken down the general problem of satellite autonomy and coordination into several sub-problems, including but not limited to retrieval and repair scenarios, satellite tasking, earth observation, space observation, formation flight and science scenarios\cite{araguz2018applying}. Use of indirect coordination methods, like Ant Colony Optimization algorithms or stigmergy methods, and direct coordination like in the negotiation scheme used in the NODES mission, have been explored for spacecraft coordination\cite{hanson2016nodes}. A common thread, regardless of the coordination plan, is to use learning or optimization to improve coordination over time. Among the research into spacecraft autonomy, which is sometimes separate from spacecraft coordination, several schema have been proposed to separate the various necessary on-board processing into parts. Most of these schema separate on-board processing into a reactive layer, which takes sensor input and reacts in low level ways, and a reflective layer, which takes information from the reactive layer and self reflects on performance. This reflective layer adjusts the reactive layer to achieve better performance towards high level goals. Across the literature, this seems to be the simplest autonomy schema that others are more complicated versions of. 

\subsection{Advantages of Learned Approaches}
The work presented in this paper is part of a larger effort to approach the problem of Satellite ADC with simulation and optimization, rather than through the development of robust (meaning error resistant in this context) deterministic algorithms. Advances in machine learning and simulation over the last 20 years mean that the space of possible pathways to satellite autonomy are more expansive than they were during the development of \textit{Deep Space 1}. It is now possible for a satellite to carry out numerous simulations onboard to test potential actions it may take. The development of learning methods based on achieving a goal, including reward based reinforcement learning or fitness based evolutionary algorithms, allow for a learned approach based on simulated or real outcomes as compared to traditional supervised learning based approaches that require a training dataset representative of the problem domain\cite{Corns_2017}. This is ideal for autonomy, which is defined by independent accomplishment of goals. So at this time, the alignment between Satellite ADC and learning methods certainly is present. The advantages of this approach, as opposed to developing robust algorithms to handle tasks autonomously, are manifold. Robust algorithms are developed by humans, and don't flexibly adapt to the problem domain. Learning from simulation enables automatic accommodation of circumstances which are not foreseen by a human developer, but do occur within simulation. Then as simulation becomes more advanced, and the problem domain more complex, it becomes more important that control algorithms learn from simulation. Robust algorithms also do not generalize, they are task specific, whereas learned approaches do generalize and can be adjusted to any task provided the reward or fitness functions are updated. Due to the reasons enumerated above, general application of learning systems, through the use of various simulation and optimization methods, promises significant impact in the problem of Satellite ADC.

\subsection{Evolutionary Algorithm Preliminaries}
An evolutionary algorithm is a type of optimization algorithm that is inspired by biological evolution. Rather than "learn" a good solution to a given problem through a system of well defined rewards and adjustments as is the case in reinforcement learning approaches, an evolutionary algorithm creates a population of candidate. These solutions "mate" with each other in some way to produce more solutions. This mating method must preserve some important aspects of the parent solutions, while also allowing the offspring solutions to differ from their parents to adequately search the solution space. Solutions have some fitness defined by the goal of the algorithm. In order for better solutions to be born, solutions with greater fitness must mate with candidates with lower fitness. This can be accomplished with either strong selection, where the less fit don't get to mate at all, or weak selection, where the less fit simply have a lower probability or proportion of mating. This is the basic form of an evolutionary algorithm, there has been significant work investigating the use of additional algorithmic steps and processes to prevent population stagnation and improve convergence\cite{Corns_2017}. We outline some of these extensions to basic evolutionary algorithms in the experimental descriptions below.

\section{Genetic Algorithm Implementation}

\subsection{Goal Task Description}
The goal was for a simulated spacecraft, placed in some starting orbit, to perform a series of thirty delta-v impulse maneuvers, or instantaneous changes in velocity, into a polar, circular orbit. The number 30 for the size of the series was chosen as a baseline, and the number of delta-v maneuvers allowed would realistically be determined based on spacecraft parameters. These parameters may include factors like onboard fuel, orbital position, and target orbit. The goal of the program was to implement a genetic algorithm which, not provided with any special start series of maneuvers, could produce a series of maneuvers which resulted in a highly polar and highly circular final orbit. Out of the six Keplerian orbit elements, the only ones considered in this experiment were eccentricity and inclination.

\subsection{Experimental Setup}
The experiment was carried out using the Poliastro Python library, an open source orbital dynamics library. Four simulated spacecraft were placed into different, non polar and non circular orbits, as can be seen in Figure 1. The size of the delta-v maneuvers they could do was constrained, and the directions were constrained to the six Cartesian directions, positive and negative x, y and z. They would perform one delta-v maneuver every 20 minutes, between maneuvers their orbit was propagated 20 minutes. At every 20 minute interval, they could also remain idle instead of performing a maneuver.

\begin{figure}
    \centering
    \includegraphics[scale=0.4]{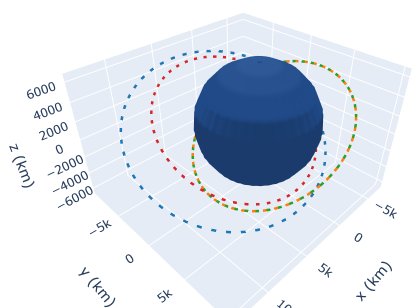}
    \caption{Initial orbits of simulated spacecraft. Varying orbital radii and starting points were selected. The plot is three dimensional, so an angle that showed the orbits best was selected.}
    \label{fig:1}
\end{figure}

\subsection{Basic Algorithm}
The basic form of the implementation of this genetic algorithm included four parts. Devising a fitness score, generating a population of solutions, mating these solutions, and populating a new generation with their children. The fitness score for this experiment was composed of two factors, the eccentricity and the inclination. The inclination of a polar orbit is 90 degrees, so the fitness score was the distance of the inclination of the final orbit from 90 degrees summed with the eccentricity of the final orbit. The final orbit in this case refers to the resultant orbit of a particular maneuver sequence. The lower the fitness score, the more fit the maneuver sequence. The population generation was performed by using a random number generator to generate a random number between 0 and 6 and creating a 30 digit long string of these numbers. A 0 meant remain idle, and each number signified an impulse in the positive x, y, z or negative x, y and z directions respectively. The mating scheme for two impulse strings was devised as the following. Element for element, if the mother and father strings agree on an element, the offspring has that element as well. If they differ, a random maneuver is put in the place of that element. This scheme was selected because a scheme was needed which would allow traits of the parent solutions to be preserved in the children while still allowing for some variance for population diversity. Finally, the next generation was populated with children solutions. For evolution towards fitter solutions to occur, the fittest solutions needed to mate the most, and the least fit needed to mate the least. Evolution was carried out over 200 generations, and the fittest solutions were output as the algorithms chosen solution. To combat stagnation in fitness score, mutation and crossover were introduced. A chance of mutation was introduced such that when two sequences mated, even if they agreed on an element, there was a small chance of the element being randomly generated. To introduce crossover, a second population evolving alongside the first was created. Every five generations, these two populations would interbreed, and their traits would crossover. When the fitness scores of both populations would stagnate, a randomly generated but seeded "immigrant" population would be brought in to mate with one of the populations. Seeded random generation means that populations were generated around an input sequence, and would have a chance of preserving the elements of the input sequence. So an "immigrant" population would be similar to the input sequence, in this case the fittest of one of the populations.

\subsection{Results}
The four spacecraft that were started in different orbits all achieved final orbits with fitness scores of less than 0.1, meaning the distance in radians from a polar inclination summed with the eccentricity was less than 0.1. The results can be seen in the Figure 2, so in conclusion the genetic algorithm developed was successful at planning a series of thirty impulse maneuvers that would place the spacecraft into polar, circular orbits.
\begin{figure}
    \centering
    \includegraphics[scale=0.4]{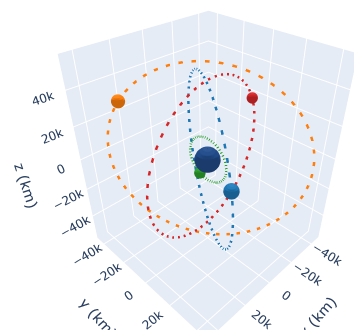}
    \caption{Final orbits of simulated spacecraft. The only orbital parameters being selected for were inclination and eccentricity, so the orbits are polar and circular but not on the same plane or the same size. The plot is three dimensional, so an angle that showed the orbits best was selected.}
    \label{fig:2}
\end{figure}

\begin{figure}
    \centering
    \includegraphics[scale=0.4]{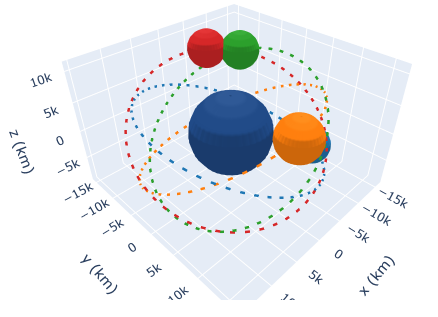}
    \caption{Final orbits of simulated spacecraft. The orbital parameters being selected for were semi-major axis and eccentricity, so the orbits are of the same size and all circular, but not co-planar.}
    \label{fig:3}
\end{figure}

\begin{figure}
    \centering
    \includegraphics[scale=0.4]{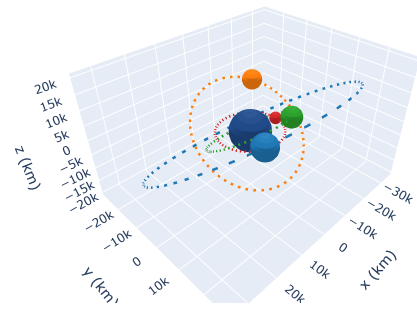}
    \caption{Final orbits of simulated spacecraft. The orbital parameters being selected for were longitude of ascending node and eccentricity, so the orbits are circular and the right ascension of the point where the orbital plane crosses the equator is zero radians, but they are not the same size.}
    \label{fig:4}
\end{figure}

Modifications to the fitness function, and nothing else, enabled the simulated spacecraft to achieve orbits close to selected values for other pairs of orbital parameters. The final orbits when the fitness was how close the eccentricity was to zero and the semi major axis was to 15000 km are shown in Figure 3, and the final orbits when the fitness was how close the eccentricity was to zero and the longitude of ascending node was to zero are shown in Figure 4. The rates of convergence of the solutions were largely the same for all the orbital parameters selected, the same number of total generations were used for each and the final solutions had similar fitness scores. Initial convergence is faster than the rate of convergence late in the process, and this is to be expected. Initial populations have high population diversity, and simple changes can drastically affect fitness in a positive way. Later in the process, simple changes can still drastically affect fitness, but usually not for the better.

\section{CONCLUSIONS AND FUTURE WORK}
In this work, we present a genetic algorithm capable of selecting a series of thirty impulse maneuvers for a satellite to achieve a desired orbit. The success of this optimization approach demonstrates a first step in the broader direction of leveraging learning systems in the coordination and autonomous direction of individual and distributed satellite systems. Future work in this line of inquiry includes achieving environmental interaction, more complex maneuvering, cooperation and coordination, multiple sequential goals, and time dependent goals of agents in an increasingly sophisticated and realistic simulated space environment using learned methods. We intend on exploring these areas in upcoming work.

\end{document}